# A Data-driven and multi-agent decision support system for time slot management at container terminals: A case study for the Port of Rotterdam

Ali Nadi[a*]

Maaike Snelder[a]

J.W.C. van Lint[a]

Lóránt Tavasszy[a]

[a] Delft University of Technology, Delft, The Netherlands

* Corresponding author: a.nadinajafabadi@tudelft.nl
2023

## ABSTRACT

Controlling the departure time of the trucks from a container hub is important to both the traffic and the logistics systems. This, however, requires an intelligent decision support system that can control and manage truck arrival times at terminal gates. This paper introduces an integrated model that can be used to understand, predict, and control logistics and traffic interactions in the port-hinterland ecosystem. This approach is context-aware and makes use of big historical data to predict system states and apply control policies accordingly, on truck inflow and outflow. The control policies ensure multiple stakeholders' satisfaction including those of trucking companies, terminal operators, and road traffic agencies. The proposed method consists of five integrated modules orchestrated to systematically steer truckers toward choosing those time slots that are expected to result in lower gate waiting times and more cost-effective schedules. The simulation is supported by real-world data and shows that significant gains can be obtained in the system.

# 1. Introduction

The problem of high waiting times for trucks at seaport terminals is receiving increasing attention from practitioners and researchers. The growing volume of international trade has put ports and their ecosystem under pressure. Long queues of idling trucks at terminal gates waiting to pick up or deliver a container lead to congestion further upstream, and induce emissions, costs, and delays (van Asperen et al., 2013, Sharif et al., 2011). Therefore, effective traffic management policies at terminal gates and the surrounding areas are becoming an imperative task for most large container ports. The problem of congestion, high waiting time, and therefore non-optimal turnaround time for trucks at the terminal is often due to a lack of port-hinterland alignment (Notteboom, 2009). Establishing such alignment concerns various stakeholders, and is highly related to the connectivity between port and hinterland (Notteboom, 2009, Wan et al., 2018). In general, the port-hinterland connection can be viewed from two perspectives. The first is the physical port-to-hinterland connectivity that can be improved through the expansion of physical infrastructure. Extending physical capacity takes considerable time and hence requires more long-term strategies.

The second perspective is digital connectivity, where multiple stakeholders can communicate and exchange information for better cooperation and coordination. Digital connectivity facilitates short-term as well as medium-term policies to control the demand patterns. Various studies (Sharif et al., 2011, Wibowo and Fransoo, 2021, Chen and Yang, 2010) found that there is potential for digital-connectivity-based solutions to control truck traffic demand patterns. This form of connectivity is often relatively cheap and fast to implement. Despite its advantages, there are also some barriers against improving digital connectivity. For example, the exchange of data and information has always been problematic due to privacy issues and fear of losing competitive advantages to other stakeholders. Recently, large ports around the world are developing safe and reliable data-sharing i.e. port community system (PCS) platforms to ease communication and facilitate digital connectivity. Even in the case of available safe data sharing platforms like PCS, the port community has not, in many cases, utilized these data properly due to the cumbersome process needed to transform big raw data into valuable information. These difficulties have led to relatively limited research towards digital connectivity as compared to physical connectivity. In this research, we contribute to the literature of enhancing digital connectivity by using shared data in port community systems and explore shorter-term solutions to solve day-to-day truck traffic issues at the terminals.

To reduce congestion at terminal gates, terminals have to balance the arrival time of the demand inflow with the available terminal processing capacity. This can happen through accurate time slot management at terminal gates. There are roughly three means of time slot management controlling demand inflows in which digital information plays a role. The first one is to provide real-time traffic information to facilitate more self-organized (user) optimal scheduling behavior of truck drivers and companies (Sharif et al., 2011). The challenge is, however, that the situations at terminals' gates may change rapidly due to the volatility of the demand. Therefore, providing real-time information may in some cases be counter-effective, even leading to trust deterioration in the system. The second approach is an incentive-based or charging-based scheme to spread demand across the day by providing monetary incentives to nudge scheduling behavior towards more (system) optimal decisions. Although charging-based policies like peak hour charges (Chen et al., 2011) can be, in many cases, effective for traffic mitigation, they may raise social objections. Incentive-based approaches also require sufficient funding sources for successful application.

The third approach towards more optimal scheduling is truck appointment systems (TAS) (Wibowo and Fransoo, 2021, Chen and Yang, 2010). A TAS typically uses a reservation system that allocates trucks to different time slots across the day based on the terminals' capacity to improve terminal efficiency. In the past decade, several results have been published around designing TAS, increasing its importance for future port development (Huynh et al., 2016, Abdelmagid et al., 2021). However, its design intricacies justify a deeper analysis of several aspects of the port-hinterland system (see Figure 1). This system includes multiple stakeholders with conflicting interests in terms of costs and benefits. For example, transport companies aim to decrease truck waiting times and container rehandling time. However, for terminal operation, it is necessary to balance the workload of the yard cranes (Im et al., 2021) and serve more customers. Also, the arrival of a large vessel may force the terminal operator to assign more cranes to the seaside and hence reduce the service rate on the hinterland side.

In order to portray the direct impacts of TAS on different stakeholders, the system can be divided spatially into the port and hinterland areas of concern and functionally into the logistics and traffic subsystems. Their relations are summarized in Figure 1.

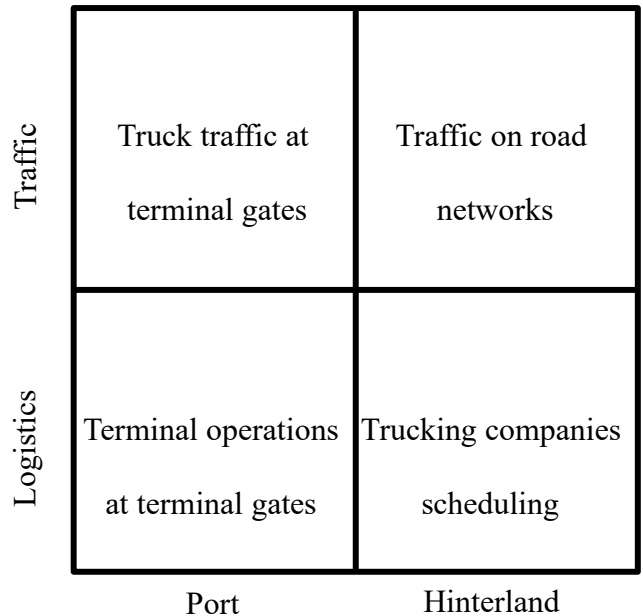

Figure 1: Connections between port-hinterland and logistics-traffic ecosystems

On the port side, the application of a TAS allows terminal operators to improve their operational efficiency. Because they can assign more cranes to the seaside operations due to reduced truck traffic at terminal gates (Chen and Yang, 2010, Zhang et al., 2013, Phan and Kim, 2015). On the hinterland side, it affects the operations of trucking companies as they may have to change their container pickup and delivery times. Additionally, changes in truck inflow and outflow could be influenced by traffic on road networks. In some cases, the peak in truck arrivals happens in off-peak periods on the road network. This could happen because truckers would like to avoid facing congestion on road networks. Although a TAS can reduce congestion at the port by shifting a portion of trucks to off-peak time slots in the port, it may add more trucks to the peak periods on the hinterland side. This not only increases the costs of truckers as they have to drive in traffic jams but also adds to the cost of other road users by increasing the percentage of trucks on road networks during congested periods. However, much of the research up to now has not treated these important traffic effects in the design of TAS. Therefore, designing a more effective TAS requires an integrated logistics, freight transport, and traffic modelling framework that coordinates between multiple stakeholders in the port and hinterland. Previous studies predominantly ignore the hinterland side while designing time slot management systems, i.e. have neglected the roadside conditions from the users' perspective. Here, we argue that multiple stakeholders in the hinterland including, trucking companies and roadside traffic agencies can also benefit or lose from the application of TAS. In this paper, we address this knowledge gap by introducing a new decision-support approach for TAS that takes all these aspects of the system into account.

The main scientific contribution of this research is the design of a comprehensive modelling framework for a TAS that integrates the different perspectives discussed above, using appropriate techniques for each problem:

1. deep neural networks that predict truck demand at port and traffic states on road networks;
2. discrete event simulation of the truck handling process at terminals' gates;
3. behavioral modelling of trucking companies' preferences for container pick-up times;
4. Mathematical optimization modelling of the time slot scheduling problem.

The model provides users of the system with the following advanced capabilities;
1. it allows priority to trucks to approach terminals' gates, based on their market preferences;
2. it provides accurate predictions of operation time at the planning horizon allowing truckers and terminals to make better decisions;
3. it suggests appropriate time slots to drivers;
4. It assures accurate coordination between multiple stakeholders in the port and the hinterland and gives a comprehensive assessment of the potential gains/costs for the application of the proposed TAS.

This paper is structured as follows: First, we review the literature for TAS. Next, we explain the modelling framework and methodology. Afterward, we present the results, and finally, we conclude the paper by discussing the findings.

## 2. Literature review

The truck appointment system (TAS) is a solution to control truck arrivals to the container terminal and improve terminal efficiency. In this system, trucks are appointed to specific time slots to load and unload their cargo considering the constraints of terminals and trucking companies. A TAS system has two main components i.e. the quota and the appointment mechanism. The quota mechanism relates to the optimum number of time slots during working hours, the duration of time slots, and the maximum number of trucks allowed per time slot (Lange et al., 2020). The appointment mechanism, on the other hand, relates to the process with which a time window is assigned to an import/export container to be picked up/delivered from/to a marine terminal. This could evenly distribute truck arrivals across the day and hence increase the efficiency of terminal operations and reduce truck waiting times at terminal gates. There is good evidence for the emergence and effectiveness of truck appointment systems to reduce congestion at seaport terminals (Giuliano and O'Brien, 2007). A recent comprehensive review of the research is given in (Abdelmagid et al., 2021). The earliest discussion on truck appointment systems emerged around early 1970 in Canada (Huynh, 2009) and soon after got considerable attention from both practitioners and researchers. In this section, we summarize the advances in TAS research, and discuss them from a methodological and a design perspective.

From a methodological perspective, researchers use mathematical programming, queueing theory, simulation, or a combination of them (Abdelmagid et al., 2021, Guan and Liu, 2009a). The mathematical programming method computes the best match between truck arrival patterns and service availability according to particular objective functions (e.g. waiting time costs, terminal handling, and emissions) and constraints. Various types of mathematical programming such as binary (M. Abdelmagid et al., 2021) and mixed-integer programming (linear and nonlinear (Xu et al., 2021)) have been utilized to solve time-slot scheduling problems in truck appointment systems. Queueing theory is very close to the physics of the system and can provide decision-makers with a good approximation of truck queue length and truck waiting times even before the real-world application of TAS (Zhang et al., 2013). Finally, discrete event simulation mimics the real-world dynamics of terminal operations in response to the application of TAS.

Building on the advantages of these methods, a large and growing body of literature has used a combination of these methods to develop a TAS. For example, M. Abdelmagid et al. (2021) proposed a binary integer programming formulation to develop a TAS that maximizes the resource utilization of the terminal while minimizing the density of trucks inside the terminal. Their approach gives higher priority to the quayside operations as compared to the truck gate operations. Guan and Liu (2009b) proposed an optimization model that minimizes various gate system costs. The cost includes labour cost, terminal operation cost, and truck waiting cost. They used queuing modelling validated with field observations to approximate these costs. Li et al. (2020) proposed a bi-objective optimization formulation to balance the trade-off between trucks and terminal operation. In their formulation, they minimize the total amount of cranes assigned to the hinterland yard blocks and the waiting time at blocks for trucks. Similarly, Mar-Ortiz et al. (2020) designed a decision support system that balances capacity management (supply) and demand management (truck arrivals). Their research aims at determining the optimum quota considering crane productivity indicators and level of service for trucks. Besides operational costs, some authors have considered emission costs while optimizing slot assignments. Do et al. (2016) developed a simulation-based genetic algorithm approach aiming at minimizing total emissions produced from trucks and cranes at import yards. They utilized discrete event simulation to estimate total truck waiting times and the total moving distance of cranes, and then minimized the associated emissions from both truck and cranes. Fan et al. (2019) also considered reducing carbon emissions while determining the optimum number of truck arrivals in each appointment period. Zhao and Goodchild (2010) evaluated the impact of arrival information on terminal rehandling and later in 2013, they proposed a hybrid queue modelling and simulation technologies to improve terminal efficiency in terms of container retrieval operation, crane productivity, and truck turn-time (Zhao and Goodchild, 2013). One year later Zehendner and Feillet (2014) involved more stakeholders from the port side in the TAS modelling. Their model not only improves the quality of service for trucks at terminal gates but also considers the crane operation for trains and vessels in a multimodal terminal.

All the above studies, although different in terms of their methodology and problem formulation, are similar in terms of the scope of the TAS design, in that they all focus on the operations of stakeholders in the port area and ignore those in the hinterland. The TAS belongs to the terminals, and trucking companies only can reserve a time slot if the slot is not fully occupied. All these methods determine the maximum number of trucks that can reserve a time slot or identify the optimum number of time slots during working hours. In recent years, a few but growing number of studies have started expanding the design of TAS towards a multi-stakeholder system. In such a design, TAS supports stakeholders to work together and optimize their objectives accordingly. (Phan and Kim, 2015) introduce a decentralized decision-making model to manage the negotiation between terminal operators and the dispatcher of trucking companies. They propose two sub-models. The first sub-model considers the scheduling of trucks based on the time window constraints posed by the terminal operator and the second sub-model approximates the waiting time based on the number of requests. This is an iterative process, which dynamically identifies the best truck schedules according to the current terminal operation. In this process, trucking companies apply for a time slot, the terminal operator approximates the waiting time for the requested time windows and sends this information to the carriers, carriers then reschedule their time slot based on the new constraint and propose a new time slot to the terminal operator. Yi et al. (2019) used a similar approach but added new constraints to the problem and proposed an efficient algorithm to solve this problem in a reasonable computation time. Torkjazi et al. (2018) expanded this collaborative method by considering tour scheduling of carriers accordingly. Here, the time slots imposed by the terminal operation influence the tour of the carrier as they are also constrained by other time windows imposed by customers. Their method helped to improve the truck tour scheduling by 11.5% compared to the model without rescheduling. The idea of a decentralized collaborative TAS has made this system more realistic and efficient. However, the drawback is that it requires a robust communications channel and a relatively long planning horizon, as trucking companies have to dynamically change their plan. This long replanning time may keep the carrier on pause as the situations at terminals can rapidly change.

In summary, most existing studies assume that trucks can follow the optimum design of the appointment system at no cost. Such truck appointment systems usually enforces truckers to choose another time slot even though this shift in their arrival may have a domino effect on their operation schedules in the hinterland. In other words, these models consider the cost that truckers would have if they had to wait in a queue but not the cost that they would have if they need to shift their arrival time due to the lack of available spots in their preferred timeslot. Only a few recent works have considered this in designing a collaborative dynamic truck appointment system. However, this requires several backs-and-forth communication, as well as negotiation between carriers and terminal operators, which significantly increases planning time. Therefore, it is essential for terminal operators to consider truckers' scheduling preferences. This helps the collaborative TAS to minimize the communication need between stakeholders. In addition, previous research so far has neglected the role of road traffic in the design of the TAS. Traffic plays a key role in the scheduling of trucks and varies according to the variation of truck demands.

In this research, we introduce a new centralized truck appointment system, integrated with a traffic model. This central TAS communicates with all stakeholders i.e. terminal operators, traffic agencies and trucking companies, and collects required information. Then, it gives accurate advice to each stakeholder according to their conditions or preferences. The focus of this research is not only on providing a normative solution for companies, as is usually the case in the literature on TAS, but also to provide a quantitative description of the system for managers and policymakers, helping them to understand the port-hinterland ecosystem better. One can think of this method as a combined logistics and traffic system where traffic at port and hinterland can be understood together and controlled simultaneously. With our method, the benefits and cost of the system can be realized and distributed in both logistics and traffic systems. This system can also deal with heterogeneity in trucking operations of multiple industries. Previous TAS studies have assumed the first-reserve-first-serve policy, and make no distinction between different preferences of shippers, e.g. between perishable agricultural merchandise with a morning delivery window and fashion products destined to warehouses in the afternoon. With the power of data-driven models deployed, our proposed method gives priority to some industries for specific time slots according to their inferred scheduling preferences.

To conclude the review, it appears that site managers at container ports require more disaggregate behavioural insights to have a better grip on demand. Data-driven methods allow exploring trucker behaviour at a disaggregated level but have, to our knowledge, never been applied to TAS. In this research, we aim to help fill this gap by introducing an optimal control policy derived from an analysis of a large historical sample of carriers and terminal gate operations. In addition, we use extensive traffic data to assure that control policies on logistics sites do not deteriorate traffic conditions on road networks. This also has never been considered in the literature.

## 3. Methodology for design of the truck appointment system

In this section, we propose a methodology for the design of a data-driven centralized time slot management system that takes multiple stakeholders into account. The building blocks of the proposed system are depicted in Figure 2.

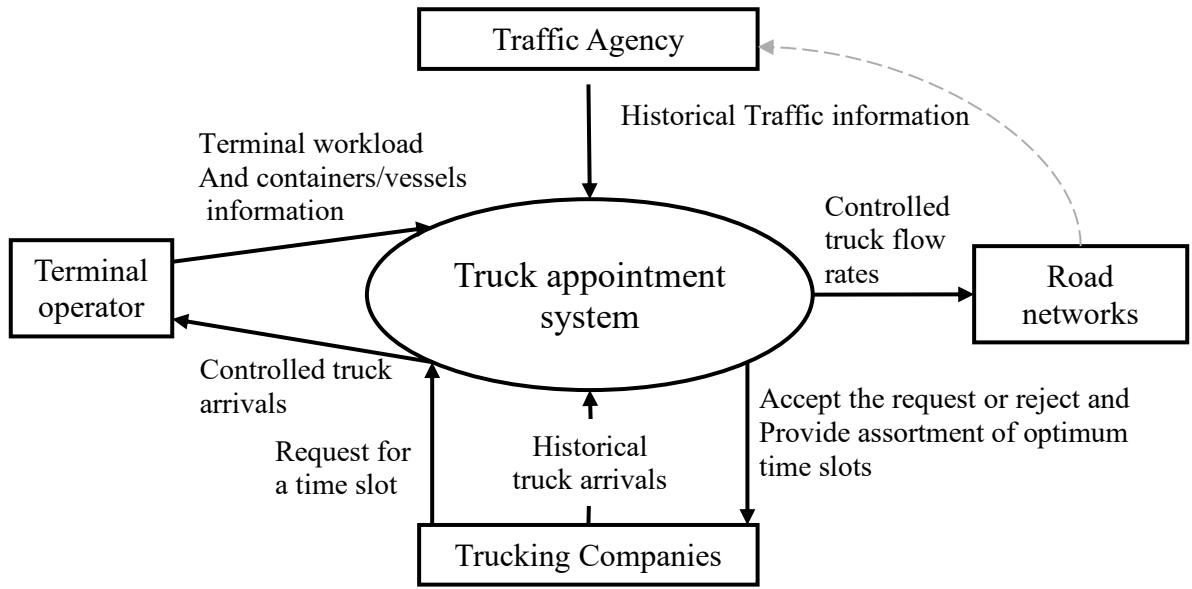

Figure 2: Centralized Multi-stakeholder truck appointment system

The process begins with carriers (trucking companies) who send a request for a time slot to the truck appointment system once they receive an order from shippers. This request may be part of a planned tour, which may include several other pickups and deliveries in the hinterland. For the TAS, however, this request is the only observable part of the tour. The TAS should learn from this partially observed tour planning information to infer the scheduling of carriers for different times of the day. On the other hand, the system looks into the terminal activities. Terminals can provide more information about their activities through the port community system. Based on this information, the TAS system can predict the workload of the terminal at the requested time slots and predict the truck waiting time cost at the terminal gate. Simultaneously, the system receives information from traffic agencies about the historical traffic conditions on road networks. This can provide an accurate prediction of delays on the surrounding road networks for trucking companies at the requested time slot. Based on this information, the system can give suggestions to each actor to help them keep their operations efficient while controlling the congestion at terminal gates. For example, it could advise the terminal operator to balance the workload of cranes between the yard side and the hinterland side based on the number of trucks approaching the gates (see Figure 3). On the other hand, if the terminal is overloaded with containers discharged from a big vessel, it may advise the terminal operator to assign more cranes to the yard operation and, at the same time, propose new time slots to carriers based on their scheduling preferences and road traffic conditions. From the traffic agency perspective, the system also can notify a traffic manager beforehand about a surge of trucks heading into the hinterland at a particular time of the day, so that they can apply appropriate traffic control measures. Our system of interest is illustrated in Figure 3.

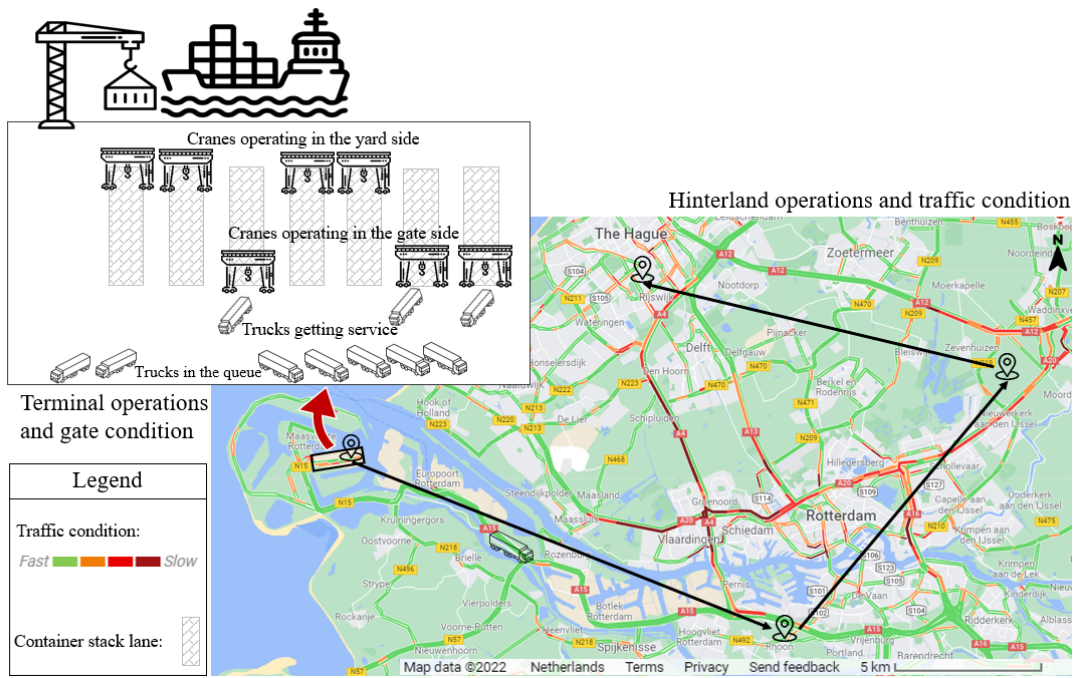

Figure 3: Logistics operations and traffic conditions in the study area

We assume that the decision time in this system begins at the end of a pre-defined planning time window (e.g. 1 hour) which can be 3 days to 3 hours before the actual operation time. The width of the planning time window is flexible and can be set by a system administrator. During this time, the system collects all the information from multiple stakeholders, and at the end of this time windows reveals the expected system states and associated prescriptions.

The objective is to minimize the total daily cost of the system which consists of three categories of costs:

1. Terminal gate operation cost $C_t^g$ which is a function of $C_t^S$, the service cost of using a crane on the gate side during time slot $t$, and $S$, the number of cranes giving service to the trucks on the gate side.

$$C_t^g = C_t^S \times S_t \qquad\qquad (1)$$

2. Carriers' costs ($C_t^C$) which includes two components:
   a. $C_t^w$: Total cost of waiting in a queue at the terminal gate during time slot t
   b. $C_t^p$: Planning or scheduling cost of visiting terminal during time slot t due to the hinterland operations
3. Societal costs of traffic systems $C_t^{tr}$ due to the traffic intensity on the main roads from and to the port

In this section, we propose a port-hinterland modelling framework considering all these costs. This framework requires 5 modules. The first module is a demand model which predicts (generates) truck demands throughout the day. The second is the container terminal gate module which includes queueing models to represent terminals gate operations. This module uses truck inflows (generated by the first module) as an input and generates truck outflows, terminal operation costs $C_t^g$ and waiting costs $C_t^w$. The third module is a data-driven traffic module that can estimate the societal cost of the traffic system $C_t^{tr}$ learning from day-to-day traffic patterns. This module uses the truck outflow generated by the second module to predict the monetary value of vehicle loss hours on the road network. The fourth module is the data-driven truck scheduling model which can predict the associated hinterland cost of trucks for picking up containers at the port in different time windows along the day i.e. $C_t^p$. The last module is an optimization model which minimizes the total system costs by assigning appropriate time slots to trucks. Figure 4 illustrates the complete modelling procedure and connection between these modules.

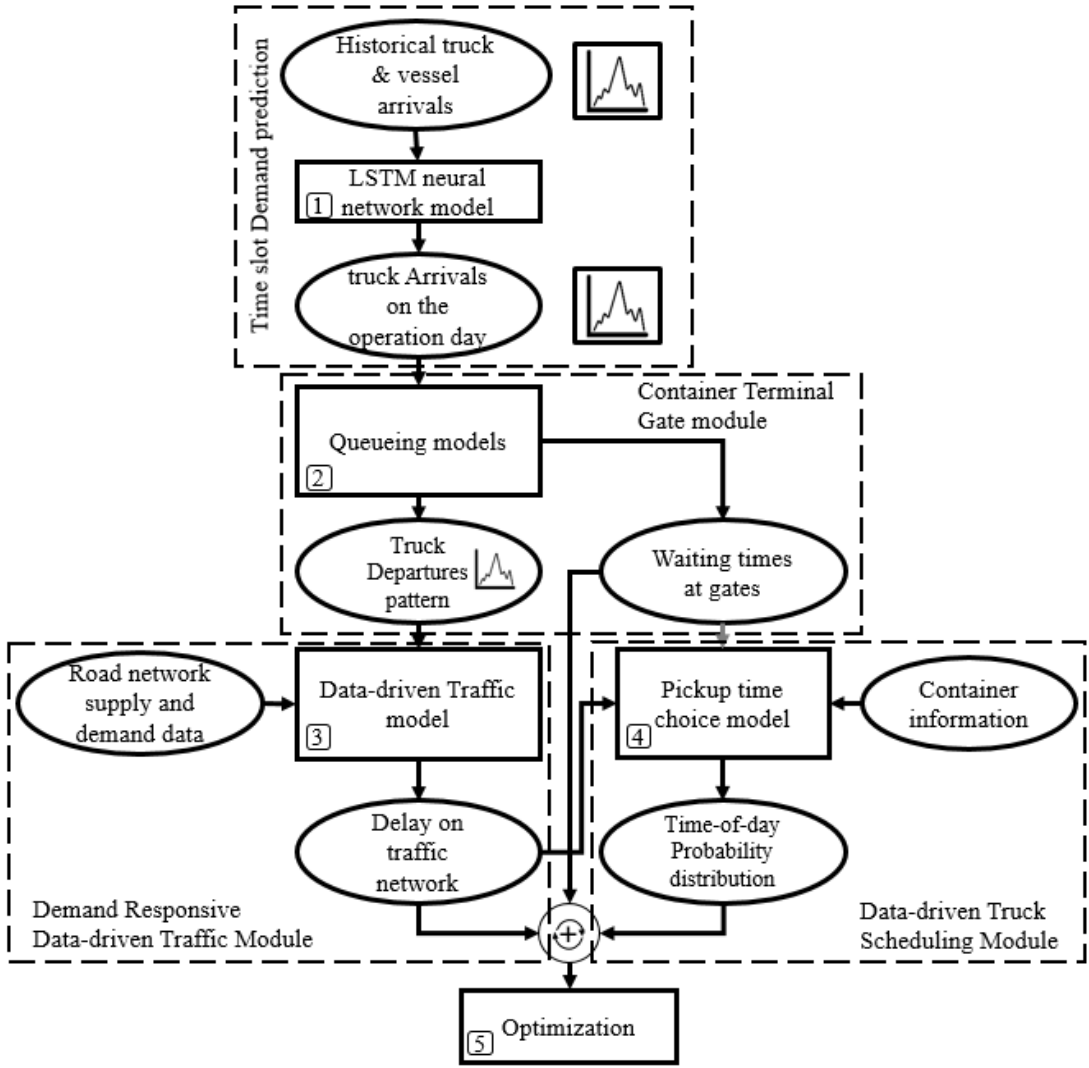

Figure 4: Modelling procedure for time-slot management system

In the next subsections, we explain the method used to develop each component as well as the data used for a case study in the port of Rotterdam.

### 3.1. Time slot demand prediction

We first divide the time horizon of the system into the planning and operation days. Assume that on the planning day at time windows $t_p$, the system receives requests for different time slots $t_o$ in the operation day. The operation day could be the same as the planning day or one to two days after that. In either case, we assume that $t_p$ is at least 3 hours before $t_o$. To predict the waiting time for each time slot on the operation day, we are first required to predict the number of trucks that arrive at each terminal. This prediction takes place on the planning day and the only information we can use for this prediction belongs to the day of planning and the days before that. To understand what information can be a good predictor of the truck demand for a time slot, we should look at the process at terminals. In a seaport terminal, once a vessel arrives, the cranes discharge containers. Then straddle carriers take the containers to the stackyard where the yard-hinterland cranes stack the container on the lanes. A container can stay in the terminal for weeks until a truck comes to the port and pick it up. Figure 5 shows an example of a typical distribution of pickup latencies, based on data from the Port of Rotterdam. The average pickup latency is approximately 4488 minutes (74.8 hours). That means it takes more than 3 days on average for containers to be picked up. We can see that the distribution is skewed and has a wide right-side tail. This buffer time depends on the container discharge time, administration time, stacking time, and carriers planning latency which may differ from container to container and firm to firm. Since not all this information is available to the system at the time of prediction, we bypass the detailed process and predict truck demands for each time slot using a time series model.

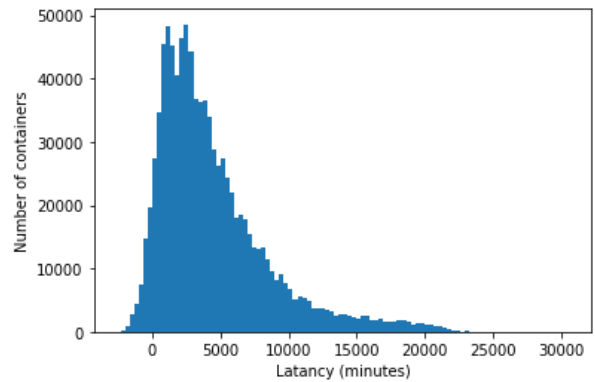

Figure 5: Distribution of pickup time  planning latency

This requires a multivariate time series that can predict truck demand based on multiple input signals. The first input signal is the historical time series of the number of road freight containers arrived by vessels at the terminal and the second signal is the time series of truck arrivals, which has autocorrelation and can deal with the recurrent daily pattern of truck arrivals.

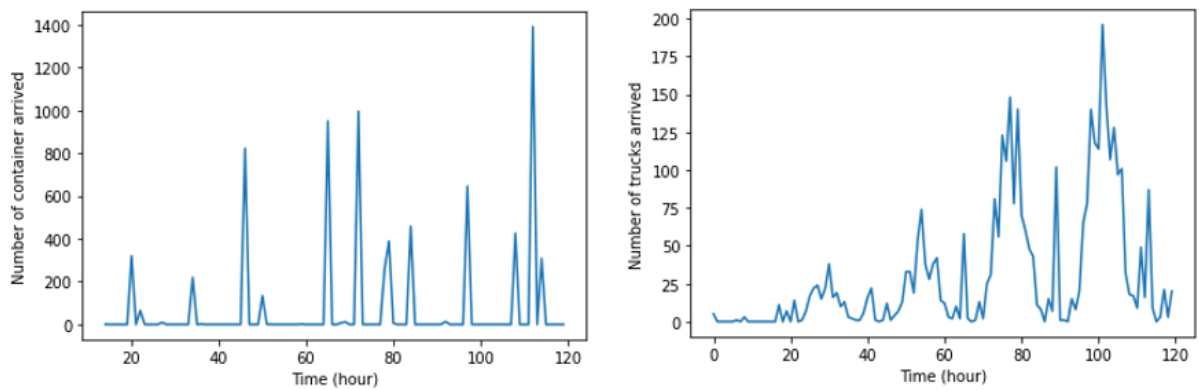

Figure 6: Comparison between deep-sea container arrival signal and truck arrival signal

Looking at Figure 6, we can see that there is a clear distinction between container and truck arrival signals. the distinction relates to the fact that the hundreds of containers arrive at the terminal all at once at a particular time of day during the call of a vessel. In other words, the deep-sea container arrival signal is zero unless a vessel arrives at the port and then the signal spikes. These containers are then distributed across different time slots in the following days as trucks arrive gradually over the day to pick them up. Since several points in the deep-sea container arrival signal contribute to multiple points in the future truck arrival signal, it is not straightforward to use this signal regularly to predict only one point in time. Instead, we formulate this problem as a multivariate multi-step time series prediction model. Equation (2) shows a general representation of this problem.

$$
\begin{bmatrix} y_{t_o} \\ \vdots \\ y_{t_o+k} \end{bmatrix} = f\left( \begin{pmatrix} x_{1,t_p-l} & \cdots & x_{1,t_p} \\ \vdots & \ddots & \vdots \\ x_{n,t_p-l} & \cdots & x_{n,t_p} \end{pmatrix} \right)
\tag{2}
$$

Our model aims at forecasting the next k values (e.g. k=24 hours) of truck arrivals time series in operation day $t_o$ from two input time series (n=2) of deep-sea container arrivals and truck arrivals in the planning $t_p$ with history lookup size of $l$ days before the planning day. To deal with this problem we propose a sequence to sequence deep recurrent neural network (RNN) for multivariate multi-step time series prediction, where we can predict the truck arrivals at all time slots of the operation day based on the combined information sequence (i.e. both deep-sea arrivals and truck arrivals) of the past days. Sequence-to-sequence deep neural network architectures have seen successful applications in various multivariate time series prediction problems (Du et al., 2020, Du et al., 2018), but not yet in our context. Figure 7 depicts a graphical illustration of the proposed model.

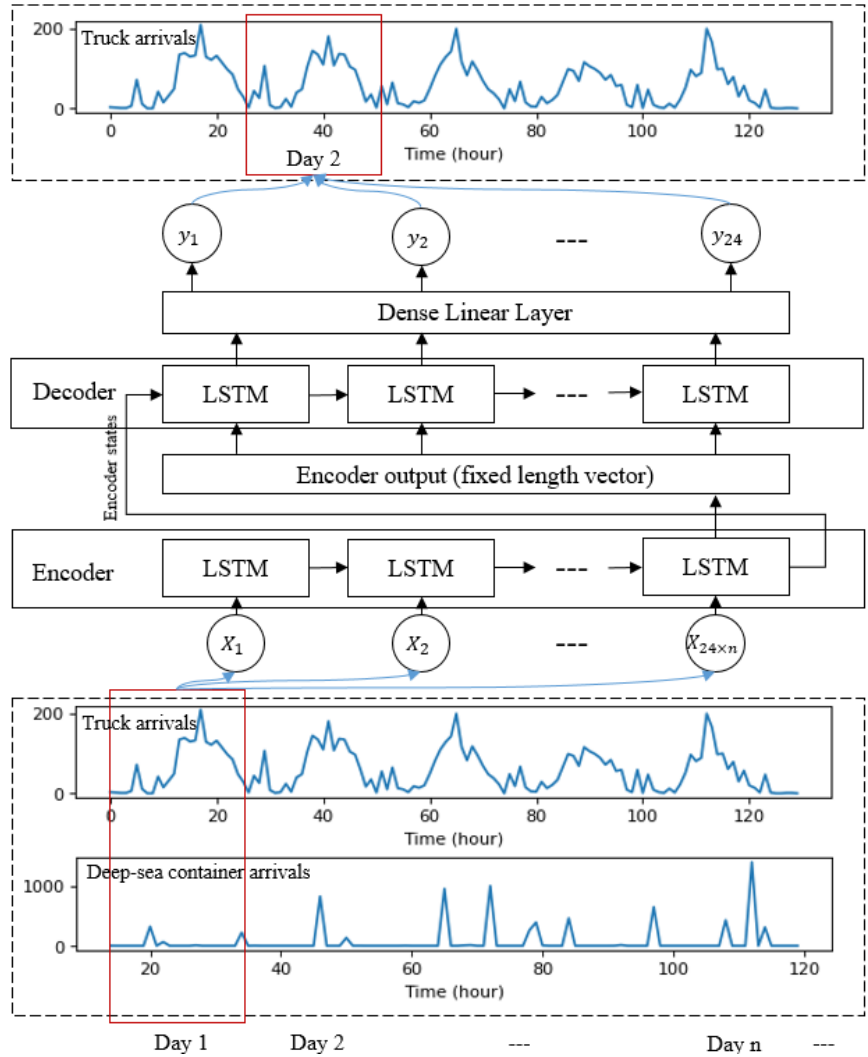

Figure 7: Architecture of the sequence to sequence deep RNN for truck arrival prediction

This model consists of two-three layers: The first layer is the encoder with several long-short term memory (LSTM) units which takes the historical multivariate time series samples as input to capture the temporal representation (fixed-length vector) of the past time series input. The second layer is an LSTM decoder, which generates the future time series as forecasting output. We used the fixed-length vector as the input and the encoder's final state as the initial state of the decoder layer. Finally, we used a linear dense (fully connected neural network) layer on top of the decoder layer to predict the target values for each time slot in the target sequence. LSTM is a type of recurrent neural network which was first introduced by Hochreiter and Schmidhuber (1997) and is known for its good performance in sequence to sequence prediction. LSTM was developed to deal with the vanishing gradient problem of traditional recurrent neural networks. A unit of LSTM includes 4 components, an internal memory cell $c_t$, an input gate $i_t$, a forget decision gate $f_t$, and the output decision gate $o_t$. Collaboration between these components enables the unit to learn and memorize long-term dependencies in a sequence. For example, in our case, the LSTM cell can learn when to consider or forget the impact of a spike in the deep-sea container arrival signal occurring in the past and, also, how big this impact would be. The compact representation of the equations for an LSTM with a forget gate is

$$i_t = \sigma_g(W_i x_t + U_i h_{t-1} + b_i) \tag{3}$$

$$f_t = \sigma_g(W_f x_t + U_f h_{t-1} + b_f) \tag{4}$$

$$o_t = \sigma_g(W_o x_t + U_o h_{t-1} + b_o) \tag{5}$$

$$\tilde{c}_t = \tanh(W_c x_t + U_c h_{t-1} + b_c) \tag{6}$$

$$c_t = f_t \circ c_{t-1} + i_t \circ \tilde{c}_t \tag{7}$$

$$h_t = o_t \circ \tanh(c_t) \tag{8}$$

Where $x_t$ is the input vector to the LSTM unit, $\sigma_g$ is the sigmoid activation function, $W$ and $U$ matrixes are the weights for input and recurrent links respectively. $b$ vector is the bias parameters and $\tilde{c} \in (-1,1)$ is the cell input activation vector. As we can see in Equation (7), $c_t$ is the summation of two terms. First is the element-wise multiplication of self-recurrent state $c_{t-1}$ and forget gate $f_t$ and the second term is element-wise multiplication of input gate $i_t$ and input activation vector.

In sum, the proposed model for demand prediction of time slots in the operation day is based on an encoder-decoder end-to-end sequence deep learning architecture that can predict the most probable future time series. In the next section, we explain how the predicted demand can be used as inputs to estimate truck outflows, waiting times, and costs at terminal gates.

### 3.2.  Container terminal gate module

There are two sides to a marine container terminal gate system. On one hand, we have the supply side which relates to the operations and gates and can be determined by the number of productive lanes and service rates (number of trucks being served per unit of time). On the other hand, we have the demand side which is the process of truck arrivals. Given the physics of the system, we can formulate the terminal gate as a multi-server queueing system with a non-stationary average truck arrival rate ($\lambda_t$), the average gate service time of trucks ($\mu$), number of active cranes at the hinterland side ($S$), average truck waiting time ($T_t$), and average number of trucks waiting ($N_t$) at time slot $t$. According to the characteristics of terminal gates, we assume that both the inter-arrival time and the service time are independent and identically distributed random variables with an exponential distribution and gates serve trucks with an integer number of cranes i.e. servers. Therefore, our method is the M/M/S queueing model. To calculate the waiting time at each time of the day we can use discrete event simulation software or the approximation method in Equation (9) developed by Cosmetatos (1976) and used in several similar studies (Guan and Liu, 2009b).

$$T_t = \{\frac{a_t^{S}}{\mu_t(S-1)!(S-a_t)^2}[\sum_{n=0}^{S-1}\frac{a_t^{n}}{n!}+\frac{a_t^{S}}{(S-1)!(S-a_t)}]^{-1} \tag{9}$$

where $a_t = \frac{\lambda_t}{\mu_t}$ denotes traffic intensity.

To set up the queue model of a terminal gate, the mean service time $\mu_t$ and the number of servers $S_t$ have to be known. In our case, this information is missing. Therefore, we estimate these two parameters from field measurements by iterative examination of different settings for the queue simulation while minimizing the mean squared differences between the simulated and observed departure profiles. After calibration, we can use the model with estimated parameters to approximate the waiting time of trucks, the number of trucks waiting at the queue, and the departure profile of trucks heading towards the hinterland all based on the predicted truck demand explained in the previous section. Using Equation (9) and Little's law, we can calculate the number of trucks waiting in the queue ($N_t = \lambda_t T_t$). Given the average truck idling cost per hour $C_I$, Equation (10) calculates the total cost of waiting in a queue at the terminal gate during time slot t:

$$C_t^{w} = C_I \times N_t \tag{10}$$

As explained in section 0, another cost of truckers in this system is the cost of visiting the terminal at the timeslot $t$ according to their tour planning in the hinterlands. In the next section, we explain how this cost can be inferred from observed data.

### 3.3. Data-driven truck scheduling module

For carriers, the cost of visiting container terminals is a part of the total cost of routing and scheduling tours along the day. Each tour may include several locations, one of which being the container terminal. The time at which a truck arrives at a container terminal depends on the sequence (in time and space) of the other pickup and drop-off locations in the hinterland. We could estimate the total hinterland operation cost of carriers for different terminal visit times if we had information about the daily activity of trucks in the hinterland. This information, however, is not observable to the system. The only observable characteristics are the containers and the arrival time of trucks to the terminal for each container. Carriers often solve a class of vehicle routing problems to plan their port and hinterland operation, minimizing their cost or maximizing their utility. Therefore, we can estimate the cost of visiting a time slot for each truck due to the relevant assumption that the utility of the truckers is maximized by their planned arrival time $t$.

$$\max \sum_{n=1}^{N}\sum_{t=1}^{T} U_{nt}T_{nt} \tag{11}$$

$$\sum_{t=1}^{T} T_{nt} = 1 \qquad \forall n \in N \tag{12}$$

$$T_{nt} = \{0,1\} \tag{13}$$

where $U_{nt}$ is the utility of truck n if visiting terminal during time windows $t$, $T_{nt}$ is a binary variable which is 1 if the truck $n$ visit terminal at time $t$ and 0 otherwise.

Equation (12) makes sure that each truck can only choose one arriving time window. We can assume that truckers use the above mathematical formulation to find the optimum visiting time $T_{nt}$ that maximizes their utility. To our system, however, the inverse of this problem matters, where the optimum vising time $T_{nt}$ is observable to the system and the system has to infer the utility of carriers $U_{nt}$. We can use utility maximization theory to calculate the probability of choosing a timeslot alternative over the others. The utility of one particular time window depends on some observed and unobserved factors that are related to the characteristics of tours and hinterland operations. We can formulate the utility function as follows:

$$U_t = V_t + \varepsilon_t \tag{14}$$

where $V_t$ is the deterministic part of the utility, which includes the observable tour attributes $\chi_i$ that influence the probability of visiting terminals in a particular time window $t$.

$$V_t = \sum \beta_i \chi_i + ASC \tag{15}$$

where $\beta_i$ are parameters of the model to be estimated from observations and capture the impact of each attribute on the utility of time windows and ASC is the alternative specific constant. The second part of the utility function contains an error term $\varepsilon_t$. This error term represents the unobserved factors that influence the utility of time windows alternatives.

We can estimate the parameters $\beta_i$ using maximum log-likelihood estimation. In the maximum log-likelihood estimation, the model aims to estimate the parameters such that the model has the highest probability of fitting the observed data. Equation (11) can be transferred to Equation (16) which presents the maximum log-likelihood function.

$$\max L(\hat{\beta}_1, ..., \hat{\beta}_k) = \sum_{n=1}^{N} \sum_{t \in T_n} T_{tn} \ln P_n(t \in T_n \mid \chi), \tag{16}$$

where L indicates the log-likelihood and $P_n(t|\chi)$ represents the probability that a trucker $n$ visits terminal during time windows t given attributes $\chi$. Using the logit function to calculate this probability will lead us to equation (17).

$$P(t \in T_n \mid \chi) = \frac{e^{V_{tn}}}{\sum_{j \in T_n} e^{V_{jn}}}, \tag{17}$$

After estimating the parameters of this model, Equation (17) generates the probability distribution of time windows for an individual truck given its attributes. This probability can relate to the utility of this truck at different time windows. We can then translate this utility as the unitless planning cost of trucker $C_t^p$ due to its hinterland operation (see section 0). In other words, if the probability of visiting terminal for a given attribute is high, it means that the cost of carriers for visiting terminal at that time windows, $C_t^p$, is low.

$$C_t^p = \eta(1 - P(t \in T_n \mid \chi))$$

(18)

where $\eta$ is a factor to scale this cost component compared to the others presented in section 0. In the next section, we explain the method to calculate the last cost component required for the design of the time slot management system.

### 3.4. Demand responsive data-driven traffic module

As explained in section 0, we also have to consider the societal cost of traffic systems attributed to the dynamics of the port-hinterland activities. Here this cost relates to the delays per km that vehicles face under different conditions of the road network. Delays can be calculated as a deviation of travel times from free-flow speed conditions. To calculate the travel time of a particular path in a network, we require the average speed of vehicles along the path. The average speed of vehicles changes in time and space and can relate to the attributes of supply and demand on a road network. For our purposes, the model should be able to assess the impact of changes in truck inflow and outflow in the port area on the traffic system and estimate the traffic-related cost of these changes. Candidate approaches are (1) macroscopic traffic models (model-driven) that use traffic flow theory to describe the spatial and temporal dynamics of principal traffic flow variables and (2) data-driven methods that use historical data to predict traffic conditions in time and space. In general, each of these methods has advantages and disadvantages and cannot outperform the other under all conditions (Calvert et al., 2015). We refer to Vlahogianni et al. (2014) and Van Lint and Van Hinsbergen (2012) for a comprehensive overview of these methods. In this paper, we opt for data-driven approaches to reduce the calibration effort and to allow daily operations with short calculation times, as well as lower computational complexity. However, we use a type of data-driven traffic model that integrates aspects of model-driven approaches from traffic flow theory to add to its transparency and interpretability.

A natural way of modelling traffic in a data-driven approach is through a non-linear mapping between demand and supply patterns. Demand concerns time-dependent inflows and outflows (summation of trucks and other vehicle classes) to the road network and supply concerns time-dependent road network characteristics like flow and speed. The correlation between the demand and supply dynamics happens in a very short-term time horizon (up to an hour ahead) (Nadi et al., 2021). However, neither demand nor supply information of operation day is observable on the planning day (often one or two days before the operation day). In previous sections, we explained how we can predict the inflow and outflow of truck demand for the operation day given the available information in the planning day and history. To estimate the impact of this predicted demand on the traffic system and calculate the traffic-related costs associated with changes in the demand, we identify two required functionalities for a traffic model. First, it should be able to predict spatial-temporal traffic states for the operation day (one or two days ahead) based on available historical traffic data on the planning day. Second, it should be able to capture the evolution of truck demand and its impact on the road network in time and space. In this paper, we skipped the first component assuming that these long-term predictions are available for the system. We, therefore, use the observed spatial-temporal traffic states of the operation day. For the real application of the system, however, we recommend the use of advanced deep neural network methods proposed for spatial-temporal long-term traffic predictions which have shown high accuracy (Zang et al., 2018).

For the second component, we need to consider temporal and spatial interaction between demand nodes (trip generation zones) and various locations on road networks. The complexity of this task is mainly there because traffic by nature is spatially correlated and highly dependent on the topology of the road network. To cope with this complexity, we use the similar traffic model that we proposed in our previous research and successfully applied it to a similar study area of our concern to assess freight departure time shift policy (Nadi et al., 2022). For the convenience of the readers, we briefly explain how this model works in this paper. This method is a graph-based convolutional recurrent neural network. The choice of this neural network architecture is also based on macroscopic traffic theory where the traffic state dynamics of a link can be described in state-space by considering the upstream link, the link under consideration, and the downstream link. In this way, the evolution of demand dynamics on the consecutive links' properties can be modeled close to the physical properties and composition of a real traffic system. This model considers the topology of the road network as a weighted directed graph $G(V, E, A)$ where $V$ is a set of nodes ( $V \in \mathbb{N}$ ) representing loop detectors on motorways, $E$ is a set of road segments in between two loop detectors and $A \in \mathbb{R}^{N \times N}$ is a weighted adjacency matrix that contains node connectivity and proximity. For this study, we aim to predict the impact of truck demand generated from the port of Rotterdam on the surrounding 5 motorways up to the boundaries of our study area.

In this model, Each node N in the graph G has a signal input of $X \in \mathbb{R}^{N \times M}$, where $M$ is the number of features of each node (e.g. volume, speed). The model also adds the signal input of the truck demand generation centroids $C \in \mathbb{R}^n$ to the passenger cars signal collected from loop detectors in the vicinity of the truck generating zones. These centroids are connected to $n \in N$ nodes in graph $G$. It is important to mention that the off-ramps and on-ramps are well connected to the other nodes to ensure that the inflow and outflow to the road network match well. Information from one node in the road network graph is passed to its neighbour nodes through the adjacency matrix. The adjacency matrix keeps the conservation law valid in the graph as the prediction of flow/speed from one node, is an input to its neighbours. To capture the spatial evolution of demand on the entire road network, The model uses the graph convolution (also known as message passing) technique. This elegant technique computes the spatial correlation of one link to its upstream or downstream links, hence it can capture spillback phenomena on a congested link (Nadi et al., 2022). In this model. First, graph coevolution is applied to the input signals and then the adjusted inputs are transferred to g which is a linear activation function of a hidden layer as presented in Equation (19).

$$f(\bar{X}, \bar{C} \mid G) = g(conv_X^* \times \theta_1 + b_1, conv_C^* \times \theta_2 + b_2) \tag{19}$$

The convolution operator $conv_X^*$ in Equation (19) calculates the aggregate (i.e. normalized weighted sum) of features for all nodes. This operator includes a particular attention mechanism and a matrix of trainable weights that produces a dynamic k-order weighted adjacency matrix for propagating each node's feature to its neighbours. With this mechanism some nodes can benefit from the information coming from their second, third, or higher-order neighbors. Since this model is demand-responsive (meaning that the graph is connected to some demand nodes generating inflows and outflows to the graph) a convolution operator $conv_C^*$ is also applied to the demand signals coming from centroids to let the demand nodes be the k[th] neighbor of all the graph nodes. Readers are encouraged to follow the original paper (Nadi et al., 2022) where there are formalisms that express how these operators work and how the model parameters are estimated in mathematical terms.

In equation 19, $f(.|G)$ is the prediction of the states in time and space on road network G. One of these states is the monetary values of vehicle loss hours regarding variations in speed and flow. In other words, inputs to the model are expected inflows and outflows, speeds, and demand signals at timestamp $t$, and outputs are speeds, flows, and monetary losses on the graph at time stamps t+Δt. Flows and speeds of vehicles are measured directly from the sensors for each link. Monetary losses are calculated based on vehicle loss hours which is the number of hours that all vehicles lose passing a link during time t as compared to if they would have passed a similar link under free-flow conditions. Equations (20) to (22) show how this monetary loss for each node $i$ at time $t$ is calculated.

$$VLH_{i,t}^{Trucks} = \max(\frac{q_{i,t}^{Trucks} l}{v_{i,t}^{Trucks}} - \frac{q_{i,t}^{Trucks} l}{FFS^{Trucks}}, 0) \tag{20}$$

$$VLH_{i,t}^{Passengers} = \max(\frac{q_{i,t}^{Passengers} l}{v_{i,t}^{all}} - \frac{q_{i,t}^{Passengers} l}{FFS^{Passengers}}, 0) \tag{21}$$

$$Loss_{i,t} = VoT^{Passengers} . VLH_{i,t}^{Passengers} + VoT^{Trucks} . VLH_{i,t}^{Trucks} \tag{22}$$

where $q_{i,t}^{passenegers}$ and $q_{i,t}^{Trucks}$ are the passenger and truck flow passing section $i$ at time $t$ derived directly from field measurements. Class-specific vehicle-loss-hours is the deviation between the flow-weighted travel time of each vehicle class in current and free flow conditions ( see Equations (20) and (21)). Finally, the *Loss* denotes the space-time monetary loss matrix of all vehicles which its elements are the summation of vehicle loss hours for passengers ($VLH^{Passengers}$) multiplied by the value of time for passengers (10 euros per hour) and vehicle loss hours for trucks ($VLH^{Truck}$) multiplied by the value of time for trucks (45 euros per hour).

The third cost component for time slot management was $C_t^{tr}$ (see section 3) which is the societal cost of the traffic system due to the changes in the arrival and departure of trucks at terminals. Given the traffic mode introduced above, we can now compute this cost for each time slot as follows:

$$C_t^{tr} = \sum_{i=1}^{N} Loss_{i,t} \tag{23}$$

## 4. Mathematical formulation and simulation-based optimization

Given all the models introduced in the above sections, we can compute all the costs in the system associated with changes in the demand and terminal service rates. In this section, we use simulation-based optimization to first simulate all these models and calculate costs and then formulate an optimization problem to manage time slots minimizing these costs. Figure 8 is a graphical representation of how this system is simulated and optimized.

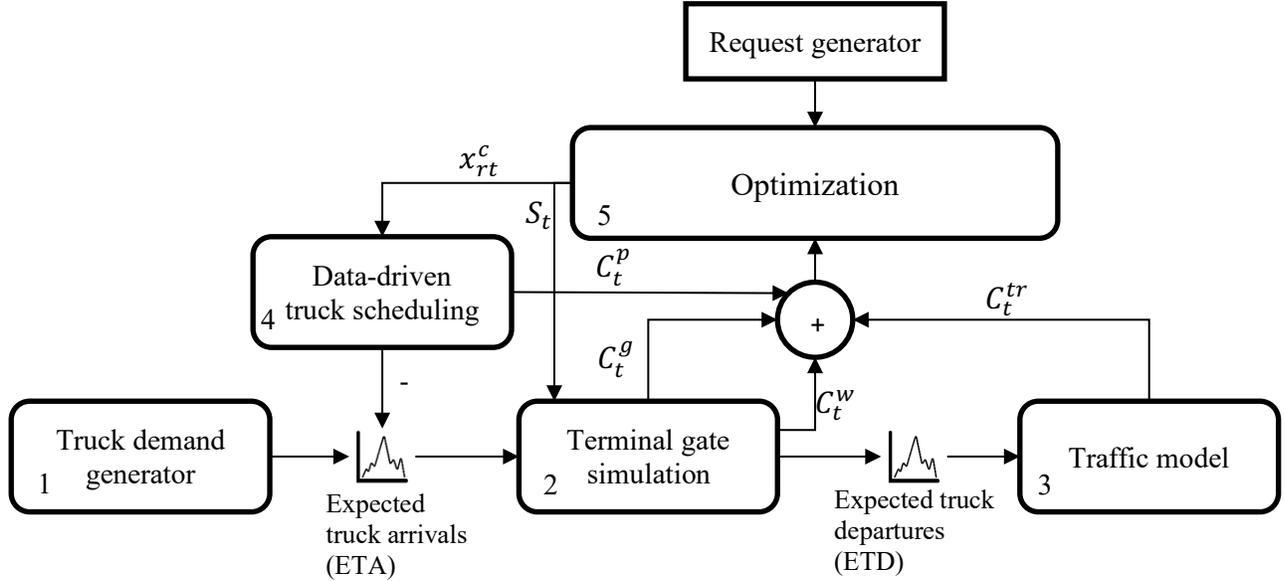

Figure 8: Simulation-based optimization framework for the time slot management system

We expand Equation 1 to formulate the optimization process in this simulation experiment. For readability and content, we present the notation of variables used in this methodology as follows:

| | |
|---|---|
| $x_{rt}$ | A binary decision variable equal to 1 if request $r$ is assigned to time window $t' \in T'$ |
| $S$ | An integer decision variable denoting the number of active cranes in the hinterland side |
| $ETA_t$ | Expected number of truck arrivals at time slot t in the operation day |
| $ETA_t^{t_p}$ | Adjusted expected number of truck arrivals at time slot $t$ in the operation day at time $t_p$ in the planning day |
| $ETD_t$ | Expected number of truck departures at time t based on the output of the queueing models |
| $R_t^{t_p}$ | A list of requests at $t_p$ in the planning day for time slot $t$ in the operation day |
| $\lambda_t$ | Average truck arrival rate at time t in the operation day |
| $\mu_t$ | Average service time for each lane at terminal gates |
| $C_t^p$ | Planning cost of visiting terminal during time slot t due to the hinterland operations |
| $C_t^w$ | Truck waiting cost per hour at terminal gates |
| $C_t^S$ | The service cost of using a crane at time slot $t$ |
| $C_t^{tr}$ | The cost traffic system due to the intensities on road networks at time slot $t$ |

This problem is a multi-objective optimization. For a given truck arrival rate (constant demand), an increasing number of active lanes (increasing supply) will increase the cost of terminal gate operation while decreasing the waiting time and vice versa. Therefore, there is a trade-off between demand and supply costs. A similar relationship also holds for traffic systems where assigning trucks to times slots with less traffic at gates may coincide with peak hours on road networks.

$$\min \ [z_1, z_2, z_3, z_4] \tag{24}$$

$$z_1 = \sum_{t \in T} \sum_{r \in R_t} C_t^p X_{rt} \tag{25}$$

$$z_2 = \sum_t C_t^w = \sum_{t \in T} f(ETA_t^{t_o}, S_t, \lambda_t, \mu_t) \tag{26}$$

$$z_3 = \sum_{t \in T} C_t^S \times S_t \tag{27}$$

$$z_4 = \sum_t C_t^{tr} \sum_{t \in T} \sum_{n \in N} loss_{nt}(ETD_t) \tag{28}$$

S.t.

$$ETA_t^{t_o} = ETA_t^{t_o - 1} - (\| R_t^{t_o - 1} \| - \sum_{r \in R_t^{t_o - 1}} X_{rt}) \tag{29}$$

$$\sum_{t \in T} ETD_t = \sum_{t \in T} ETA_t \tag{30}$$

$$\sum_{t_o \in T_o} R_t^{t_o} \leq ETA_t \tag{31}$$

$$S_t < S_{\max} \ \ \& \ S_t > 0 \ \ \& \ S_t \in \bullet \tag{32}$$

$$X_{rt} \in \{0,1\}, e_{ij} \in \{0,1\} \tag{33}$$

In the above formulation, we have five objective functions. Equations (24) to (28) belong to the planning, waiting, service, and road network traffic costs respectively. Equation (29), updates the expected number of truck arrivals recursively based on the assigned trucks in the previous planning time window. This implies that the optimization problem should be solved for each planning time window $t_o$. Equation (30) concerns the conservation of flow. Equation (31) guarantees that the cumulative number of requests for each time slot does not exceed the expected number of truck arrivals. Finally, Equations (32) and (33) show the search space and boundaries of the decision variables.

The problem presented above is a multi-objective non-linear mixed-integer optimization problem, in which the simulation optimization procedure presented in Figure 8 is used to obtain an optimal assignment of requests for each time slot, as well as an optimal value for the $S$ (number servers) to minimize the total daily cost of the port and hinterland system. This procedure is as follows:

1- First of all, a comparison between the container data and traffic data is used to convert the number of containers to the number of trucks for each time slot t in the planning day. For each planning, time window $t_p$, a Monte Carlo simulation method is used to draw and generate requests for each time slot in the operation day. These requests are proportionally selected from the observed distribution of truck pick-up time reported in the PCS data. These requests have all the information about the requested time slot, container and commodity type, and the deep-sea arrival time of the container to the terminal.

2- We use the requested time slots as the initial solution in the first iteration of the optimization process.

3- We calibrate and validate the queueing models for each terminal based on field measurements.

4- Next is estimating the data-driven truck scheduling model and simulating this model to calculate the cost (using equations (18) and (26)) of each time slot for each request generated in step 1 (attributes of requests are inputs to the truck scheduling model).

5- Build and simulate the truck generator module to generate the expected number of truck arrivals for the operation day and adjust it based on the assignment solutions (using equation (29)).

6- Use the adjusted truck arrivals and the number of active container lanes as an input to the terminal models to calculate the waiting costs and expected number of truck departures.

7- Train the traffic model as explained in section 3.4 and use the sum of the time series of the expected number of truck departures of all terminals with the time series of passenger cars. Use the result as an input to the traffic model to predict the loss hours of all vehicles on the network.

8- Finally, use a multi-objective optimization algorithm to find the Pareto frontier of solutions making a trade-off between all these costs.

We use the optimization toolbox of MATLAB R2019b as a solver to solve this problem. This toolbox uses a controlled, elitist genetic algorithm, a variant of NSGA-II (Deb et al., 2002), to find the Pareto front of solutions.

## 5. Experimental setup

As explained in the previous section and illustrated in Figure 8, our method consists of five main modules. The first module is to predict the time series of truck demand for each terminal; the second is the marine terminal gate model to use these predictions to estimate waiting times and trucks' departure time series; the third is the data-driven traffic model to predict traffic states (flow, speed, and vehicle loss hours) across the road network. We use the port of Rotterdam as our case study. The fourth is the data-driven truck scheduling module which estimates the cost of scheduling a truck to pick up a container at a particular time window within a day; and finally, In this section, we demonstrate and validate these components before being used in the simulation-based optimization procedure. We begin by introducing the available datasets and then explain and validate the parameters of all models.

## 5.1. Logistics and traffic data

Two main categories of data i.e. logistics and traffic data are used in this research for the estimation of models and their validations.

- The logistics data is twofold. One is the PCS container data (provided by the Portbase) for five terminals operating in the Port of Rotterdam. These data include the deep-sea arrival time of vessels, container discharge time, container size/type, commodity type, and the time of loading containers on trucks. This data is used to estimate the data-driven truck scheduling model as well as to generate time slot requests for the simulation. The truck demand generator also utilizes the historical deep-sea arrival data stored in this data set in combination with historical truck inflows to predict truck demand. We also use data from two trucking companies reporting on their idling time at terminal gates. This data is used to validate the marine terminal gate queuing model.

- Traffic data are collected from loop detectors located on the road network (provided by the national data warehouse NDW) and also in the vicinity of terminals gates (provided by the Port of Rotterdam). Road network data includes a stream of flow and speed data in 1-minute resolution and is used to estimate the traffic model and the gate data includes inflow and outflow to the terminal per hour which is used to calibrate marine terminals' gate models.

## 5.2. Demand model estimation

As explained in section 3.1, we use a sequence to sequence deep LSTM network with encoder-decoder architecture for multi-step forward prediction of truck demand trough out the operation day. We used the data collected for the whole year of 2017 from PCS for this model to predict hourly demand for the operation day. We used 10 months (January to October) of data for training and the months of November and December for testing and validating the model respectively. We used the Adam algorithm, 40 units for each hidden layer, the learning rate of 0.005, the dropout rate is 0.2, the validation frequency is 20, and the gradient threshold set to 1 to prevent gradient explosion. The mean square error (MSE) is used as the loss function. **Error! Reference source not found.** shows the prediction results and performance of the model on the tests set for all four terminals. Terminals are anonymously labelled A to D due to the privacy regulations of the data provider.

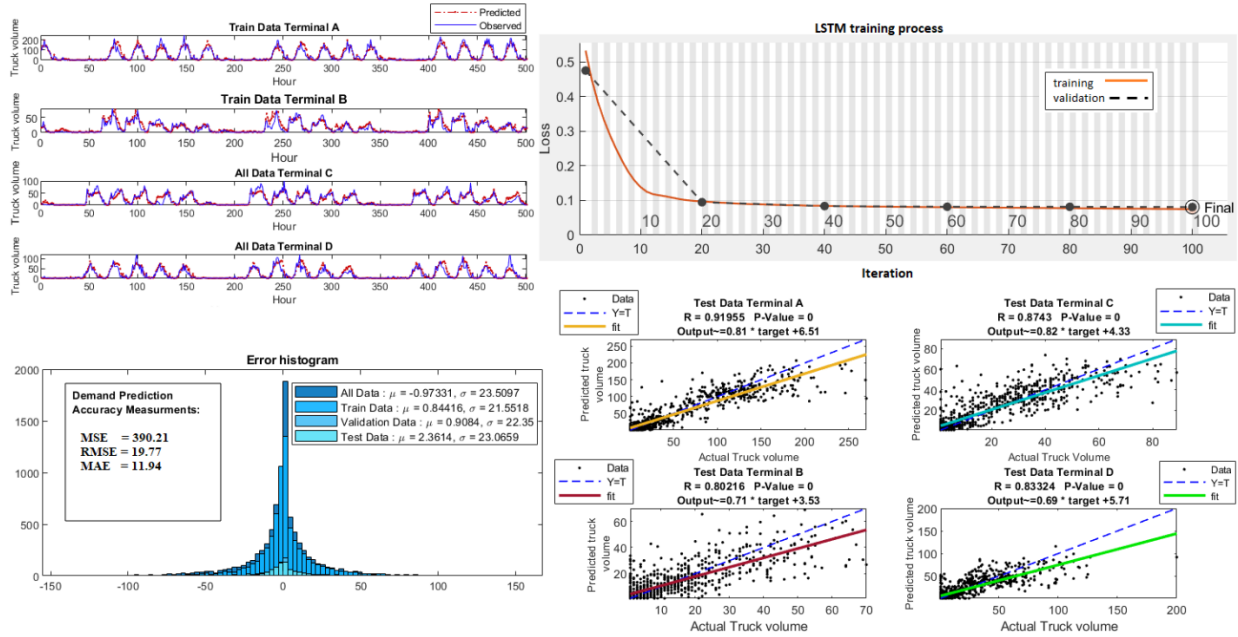

Figure 9: Performance of the demand model for 1 day ahead hourly demand prediction for each terminal

Several common accuracy measurements are used to show the performance of the model. The route means square error (RMSE) of this model is 19.77. Mean Absolute Error (MAE) is the indication of how big are the model errors on average. Figure 9 shows that the model can predict the truck arrivals with on average 11.94 error. The correlation coefficient with predictions and observed values are in all models close to 1 (between R=0.8 for terminal B and R=0.91 for terminal A) on test data. This confirms the high performance and generalization of the model to predict unseen data.

The model is tested on different lookup windows as it should be able to predict the demand depending on the day of planning. In scenario 1, the request comes on a similar day as the operation day. In this case, the lookup window will be one and two days before the operation day and a similar weekday in the last three weeks. In scenario 2, requests arrive one day before the operation day. In this scenario, the lookup time is two and three days before the operation day and similar weekdays in the last three weeks. Finally, if time slot reservation takes place two days in advance, then the step of forwarding prediction is 2 days and the lookup window will be three and four days before the operation day and the similar weekdays in the last two weeks. Table 1**Error! Reference source not found.** compares the average performance of this model for all terminals under various scenarios.

Table 1: accuracy of the model for different prediction horizons on test data

| Scenario | Lookup window | Forward steps | MSE | RMSE | MAE |
|---|---|---|---|---|---|
| 1 | $t_o - 1, t_o - 2, t_o - 7, t_o - 14, t_o - 21$ | 1 day ahead | 390.21 | 19.77 | 11.94 |
| 2 | $t_o - 2, t_o - 3, t_o - 7, t_o - 14, t_o - 21$ | 2 days ahead | 481.7 | 21.94 | 13.03 |
| 3 | $t_o - 3, t_o - 4, t_o - 7, t_o - 14, t_o - 21$ | 3 days ahead | 483.25 | 21.98 | 13.38 |

A comparison (see Table 2) between the performance of this model (for 1 day ahead) with historical average and regular LSTM time series shows that long-term demand predictions can be improved by sequence to sequence deep neural network structure.

Table 2: comparison of the model performance with other time series models

| model | MSE | RMSE | MAE |
|---|---|---|---|
| Seq2Seq LSTM | 390.21 | 19.77 | 11.94 |
| Regular LSTM | 450.42 | 21.22 | 12.8 |
| HA | 604.39 | 24.58 | 14.96 |

### 5.3. Marine Terminal gate model calibration and validation

As explained in section 3.2, the terminals' gate models are parametric. These parameters have to be determined with field measurements to ensure that the models can simulate the gate operations close to an average day in the real world. These parameters are $\lambda_t$, $\mu_t$, and S which are truck arrival rate, gate average service rate, and the number of servers (lanes). Please note that the service time of the servers is generated using a constrained exponential distribution with service rate $\mu_t$ which will not generate service times less than 20 minutes. We calibrate four terminal models based on an average daily profile for weekdays since the statistical tests, e.g. ANOVA and two-sided t-test, showed that there are no significant monthly or daily trends in arrival or departure profiles. We have used the method of least squares to estimate these parameters, based on observed average truck arrivals and departures. For terminal five, we did not have truck count observations in the vicinity of its gates. Therefore, we cannot calibrate a queueing model for this terminal. We, therefore, scaled the aggregated result of the four calibrated terminal models using loop detector data at the beginning of the road network before using it as an input in the traffic model. Table 3 shows that the simulated truck departure volumes fit the observed volumes with R-squares greater than 0.91 for all terminals.

After parameter estimation, we used a two-sided t-test to statistically test the similarity between distribution of simulated and observed truck arrival signals as well as truck departure signals. The test results show that the null hypothesis -The observed and simulated departure profiles are similar- is accepted with t-values 0.01, 0.014, 0.025, 0.029 for terminals A ,B, C, and D respectively.

Table 3: Terminal model calibration and validaton results

| Terminal | t-values | R-squared |
|---|---|---|
| A | 0.01 | 0.92 |
| B | 0.014 | 0.95 |
| C | 0.025 | 0.91 |
| D | 0.029 | 0.97 |

### 5.4. Data-driven Truck scheduling model estimation

In this section, we explain the estimation process of the parameters of equation (15) and the results of this model. Assumptions about the distribution error term result in different model specifications. The simplest way of parameter estimation for such a model is restricting all convinces to be zero meaning that all the variables should be independent and identically distributed. This results in a multinomial logit model.

Table 4 shows the overview of attributes that we consider in this study to predict the utility of pick-up time windows. Since this model is one of the first in its kind, we selected these attributes based on expert knowledge at the port of Rotterdam and reviewing the most relevant studies (Irannezhad et al., 2019, Kourounioti and Polydoropoulou, 2018, Kourounioti et al., 2021). We believe the characteristics of containers like their type, size, and weight may logically influence the utility as some of these containers, e.g. refrigerated containers, are specialized for specific use at a particular time of day. Commodity type also represents the industry that uses these commodities in the hinterland which may have specific regulations in the time of day. Traffic conditions on surrounding road networks are another attribute that most logically influences the utility of time slots for carriers. Carriers often have some expectations about the traffic conditions at different times of the day. We, therefore, incorporate these expectations in our model using average inbound and outbound delays per km per time of day that truck drivers may experience approaching the port or heading towards the hinterland. Finally, we noticed that a noticeable number of carriers come to pick up the containers on the same day as the vessel arrives. Due to the discharge process time and container administration time, there is always some latency between the arrival of the vessels and the arrival of the trucks on the same day; therefore this could influence the utility of carriers for a particular time of the day.

The alternatives in our model are times of the day which are aggregated into four-time windows. The periods are formulated as morning (from 5:00 until 10:00), midday (from 10:00 until 15:00), and afternoon (from 15:00 until 19:00), night (from 21:00 until 5:00). These periods are based on observed arrival patterns and time slot categories used in practice at the terminals.

Table 4: Overview of attributes

| Attributes | Description |
| --- | --- |
| Container type | GP: General purpose containers |
|  | RE: refrigerated containers |
|  | CC: chemical contained containers |
|  | TC: Tank containers |
| Container Length | 40ft: containers with 40 feet length |
|  | 20ft: containers with 20 feet length |
| Container weight | Categorized into: |
|  | Heavy: 15000 to 35000 kg |
|  | Light:   2000 to 15000 kg |
|  | Empty: < 2000 kg |
| Commodity type | AGR: Agricultural |
|  | Chem: Chemical products |
|  | Food: Food products |
|                   | Fert: Fertilizers                                                        |
|                   | Pet: Petroleum                                                           |
|                   | RawMin : Raw minerals                                                    |
|                   | SolMin: Solid mineral fuels                                             |
|                   | Ores                                                                     |
|                   | Miss: miscellaneous                                                      |
| Vessels size      | Large: if call size > 1250                                               |
|                   | Small: if call size < 1250                                               |
| Traffic conditions| Delay_Port: Average delay per km toward Port                             |
|                   | Delay_Hint: Average delay per km toward Hinterland                       |
| Planning latency  | The differences between arrival of the vessels and arrival of trucks if that happens to be within similar days. |

We present the result of this model in Table 5 including the model fit, estimates of the coefficients, and the level of significance (t-value). In this table, ASC represents the alternative specific coefficient which captures the mean on observed preferences for a specific alternative and β indicates the estimated coefficient for each attribute. The label of alternative-specific attributes is followed by the suffix, Mor, Mid, Aft, or Night. We have run several experiments with different specifications to get the best model fit with significant improvements in the likelihood ratio.

Table 5: Results of the truck scheduling model for all terminals (the base alternative is Night)

| Coefficient | Morning | | Midday | | Afternoon | |
|---|---|---|---|---|---|---|
| | Estimate | t | Estimate | t | Estimate | t |
| $ASC_{Night}$ | -0.251 | -12.2 | -0.251 | -12.2 | | |
| $\beta_{Agr}$ | 0.321 | 10.2 | -0.128 | -5.6 | -0.147 | -8.57 |
| $\beta_{Chem}$ | -0.101 | -4.89 | 0.111 | 7.84 | | |
| $\beta_{Fert}$ | | | 0.295 | 15.3 | 0.232 | 7.68 |
| $\beta_{Food}$ | | | 0.411 | 14.3 | 0.275 | 7.11 |
| $\beta_{Iron}$ | | | 0.278 | 5.88 | 0.356 | 6.36 |
| $\beta_{Miss}$ | | | 0.2 | 11.8 | 0.161 | 5.85 |
| $\beta_{Ores}$ | 0.144 | 5.04 | 0.263 | 11.8 | -0.166 | -5.31 |
| $\beta_{Petro}$ | | | 0.197 | 10.1 | 0.089 | 3.01 |
| $\beta_{RawMin}$ | 0.0891 | 3.26 | 0.0935 | 4.87 | | |
| $\beta_{SolMinFu}$ | -0.0838 | -3.15 | 0.2 | 10.6 | | |
| $\beta_{GP}$ | -0.292 | -14.2 | 0.312 | 18.1 | -0.0808 | -3.63 |
| $\beta_{RE}$ | -0.211 | -7.11 | 0.194 | 8.26 | -0.104 | -3.42 |
| $\beta_{CC}$ | -0.352 | -15.6 | 0.425 | 22.7 | | |
| $\beta_{TC}$ | -0.336 | -11.8 | 0.277 | 12.7 | | |
| $\beta_{Vessel\_Mor}$ | -0.233 | -5.26 | 0.385 | 10.3 | 0.53 | 13.7 |
| $\beta_{Vessel\_Mid}$ | | | 0.425 | 13.5 | 0.771 | 24.2 |
| $\beta_{Vessel\_Aft}$ | | | | | 0.619 | 17.7 |
| $\beta_{Empty}$ | 0.156 | 12.5 | 0.138 | 12.1 | 0.0979 | 8.14 |
| $\beta_{HeavyWeight}$ | 0.0638 | 4.58 | 0.359 | 28.4 | 0.0826 | 6.1 |
| $\beta_{LightWeight}$ | -0.109 | -8.1 | 0.151 | 12.2 | -0.0374 | -2.88 |
| $\beta_{Lenght\_20ft}$ | -0.0563 | -5.63 | 0.425 | 50 | 0.0758 | 7.63 |
| $\beta_{Length\_40ft}$ | -0.204 | -19.9 | 0.205 | 25.1 | -0.132 | -12.9 |
| $\beta_{Delay\_Port}$ | -0.483 | -13.9 | -0.483 | -13.9 | -0.483 | -13.9 |
| $\beta_{Delay\_Hint}$ | -1.14 | -29.7 | -1.14 | -29.7 | -1.14 | -29.7 |

Sample size: 303930
Final log-likelihood: 397581.9
Rho-squared with respect to constants: 0.19
Number of parameters: 61

Comparing the estimated coefficients of different commodity types, we can conclude that agricultural goods have a higher utility for morning time windows. This means that scheduling time slots for carriers with agricultural products will have a high cost for them in midday and afternoon. Ores and Raw mineral products also have a positive impact on the morning utility, However, their utility is slightly higher for midday. Iron is the only commodity type with the highest utility in the afternoon. The model does not suggest any significant preferences toward morning arrivals for fertilizers, foods, Iron, Miscellaneous, and petroleum which means that their preferences are more significant in the midday or the afternoon. All container types and lengths have a larger positive impact on utility for the midday. Please note the differences between the magnitude of the parameters indicated for the time slot management system to prioritize carriers based on their utilities. As opposed to the lightweight containers which have a negative impact on the utility of the morning time slots, the sign of impact for heavy containers is positive for all alternatives. However, the impact of afternoon time windows on the cost of carriers is lower than morning. Regarding planning latency, truckers that are willing to pick up a container within the same day of its arrival at the terminal have higher preferences for the next time windows. For example, if the vessel will arrive in the morning, there is strong disutility for the morning time slot for trucks - obviously because containers have to be first discharged from the vessel and only then become ready for pickup. Having this in the model helps the time slot management system give priority to these truckers to reserve a slot in the afternoon.

The sign of delay on surrounding road networks is plausible (negative). This means that the higher the delay per km, the lower the utility of the truckers. Comparing the magnitude of parameters for delays shows that the utility of carriers is more sensitive to the Hinterland direction as compared to the Port direction. It is important to mention that traffic conditions change rapidly on the road network. Therefore, we used the actual delay for the hour of the day that is associated with the pickup time of each container. In other words, delays are not aggregated for the range of alternatives. Please also note that delay is used as a generic parameter for all alternatives.

In sum, this model provides plausible estimates and will support the time slot management system to consider the utility or cost of each time window for carriers.

### 5.5. Traffic model estimation

The parameter of the proposed traffic model is estimated for the 181 days of data collected for the given road network. In Table 6, we summarize these network specifications and prediction accuracy for speed, flow and vehicle loss hours, which are then converted to the monetary loss providing traffic cost for the time slot management system. As one can see, the model can predict time-space traffic conditions accurately.

Table 6: Road network specifications and prediction accuracy

| Network description | 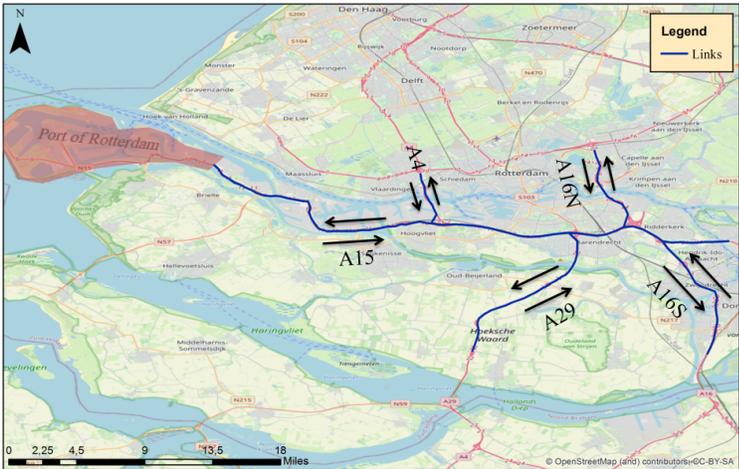 | |
|---|---|---|
| Aggregation in time | 15 min | |
| Aggregation in space | 600 m | |
| Number of nodes | 166 | |
| Path Length | A15 : 55 km ; A4: 4,8 km; A29: 11,4 km; A16 North: 7.1 km; A16 North: 20.1 km | |
| | Predictions | Observations |
| Speed | 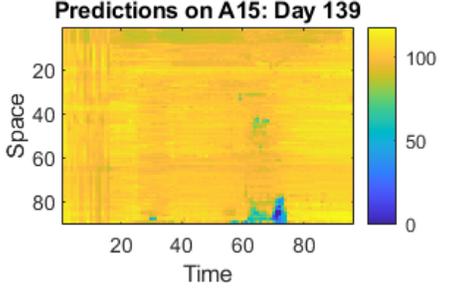 | 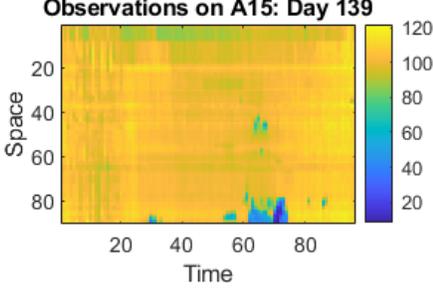 |
| Loss hours | | |

| | 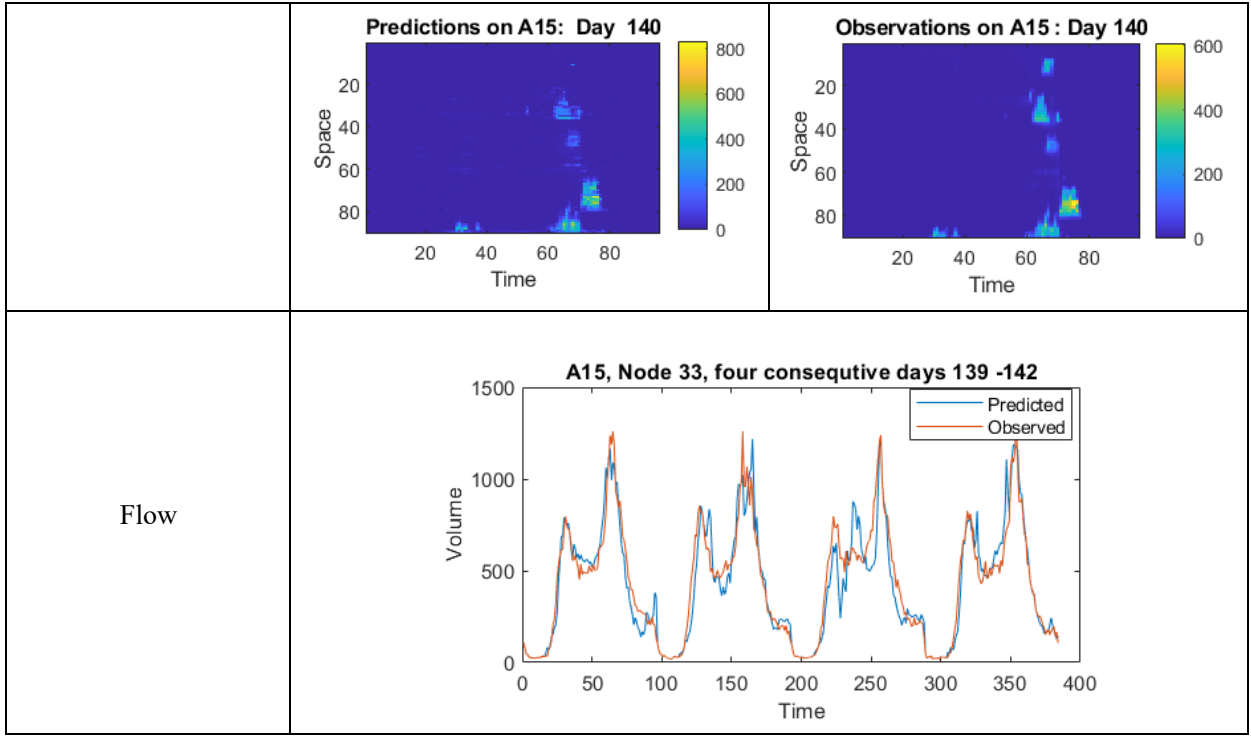 |
|---|---|
| Flow | |

We used the Mean absolute percent error (MAPE) to evaluate the accuracy of this model.

$$MAPE = \frac{1}{N} \sum_{i=1}^{N} \left| \frac{t_i - y_i}{t_i} \right| \times 100\% \tag{34}$$

where $t_i$ is ground truth and $y_i$ is the predictions. We calculated MAPE under various equally distributed error bins. This will result in a more realistic percentage of errors. the overall MAPE for speed prediction in the network is 11%, for loss hours 14%, and for flow 9%.

### 5.6. Time slot management simulation and optimization

The simulation experiment follows the steps explained in section 3. We selected a day with congestion between 15:00 and 18:00 as the simulation day. We use the container pick-up time reported in the PCS data for the selected day to generate requests. For this, we randomly but proportionally draw from the PCS data. We considered 1 hour for the time to respond on the planning day ($t_o$). The request generator generates requests for each hour between 7:00 and 17:00 which is aligned with the working hours of the trucking companies. The sum of all requests for each hour should be equal to the total demand of that hour. For each planning window, we use a MATLAB solver to re-schedule requests solving the optimization problem presented in section 3. In the year 2017, the waiting cost for trucks in container transport was estimated at around 38 euros per hour. The hourly cost of a crane is twofold. One relates to the labour cost which is equal to 2.5 people per gate lane (Guan and Liu, 2009b) and the other is the cost associated with the yard operation if the crane will be active in the hinterland side. Given 50 Euros for labour costs, we can calculate the cost of using a gate as follows:

$$C^S = 2.5 \times 50 + \alpha [C_{peak}^W(s) - C_{peak}^W(S+1)] \tag{35}$$

where $C_{peak}^W(s)$ is waiting time cost during peak hours. The idea is that the value of a crane for yard operation is at least equal to the benefit of the waiting-time reduction in the hinterland adding one more crane to serve outbound trucks. Since for the calibrated terminal models we have the estimated S for the peak hour (see section 5.3), we can now estimate this cost for the constant service and arrival rates running the terminal model with S+1. In equation (35), $\alpha > 0$ is a factor that can be tuned by terminal operators depending on the priority they may want to give to the yard operation, for instance, if large vessels are at berth. Based on our models and with $\alpha = 1$ we came up with approximately 80 Euros for the disutility of the terminal operator to add an extra crane. Therefore, we consider $C^S = 205$ Euro per hour per crane. We refer readers to equations (18) and (23) for the calculation of the re-scheduling cost for carriers changing their schedules $C_t^P$ and the cost associated with the delays on road networks $C_t^{tr}$ respectively.

Given these costs, the optimizer gives a set of solutions that are equally efficient and which together represent the Pareto frontier of search space. This Pareto frontier is difficult to visualize as it has four dimensions. The selection of one solution among all optimal solutions is straightforward, however, it depends on the policy of the decision-maker. These solutions range between an extreme focus on supply (terminal costs) and an extreme focus on demand (carriers). If the decision-maker focuses only on terminal costs (supply) and therefore selects solutions with only the lowest number of cranes as compared to the base case, this increases the waiting time for a constant arrival rate. Therefore, the optimizer reduces the arrival rate by assigning more trucks to off-peak hours which results in reduced waiting times. This, however, increases the disutility of carriers since they have to bear more re-organization costs. The cost of the traffic system changes as well due to the changes in the arrival and departure of trucks. The magnitude of the traffic cost, however, depends on the traffic conditions. In this scenario, although the solution is optimal, all the pressure will be on the carriers. In another scenario, the focus may be on the demand side (waiting time). In this approach, the system increases the number of cranes, and the waiting time drops accordingly. There are no changes needed in the arrival profile of trucks and therefore the re-organization and traffic costs will be zero in this scenario (highest carriers satisfaction). In this scenario, also the solution is optimal, but all the pressure will be on terminal operators.

In both above scenarios, only one stakeholder should take the action for the benefit of the other stakeholders in the system. Such solutions are less probable to succeed and more difficult to implement in practice since the costs of one stakeholder will be relatively high and it requires extra efforts and governance to collect a part of the gain from other stakeholders to compensate and incentivize the actor. Due to these difficulties, the most logical way is to select solutions that involve all stakeholders in the operation and balance out their costs and benefits. Among the solutions that satisfy this requirement, we select the one with the maximum monetary gain by comparing the waiting times before and after applying time slot management. The simulation result of TSMS with respect to the reduction in waiting time is illustrated in Figure 10. We can see that the waiting times are spread across other times of the day resulting in extra waiting times during the morning. The overall state of the system, however, obtains significant improvements.

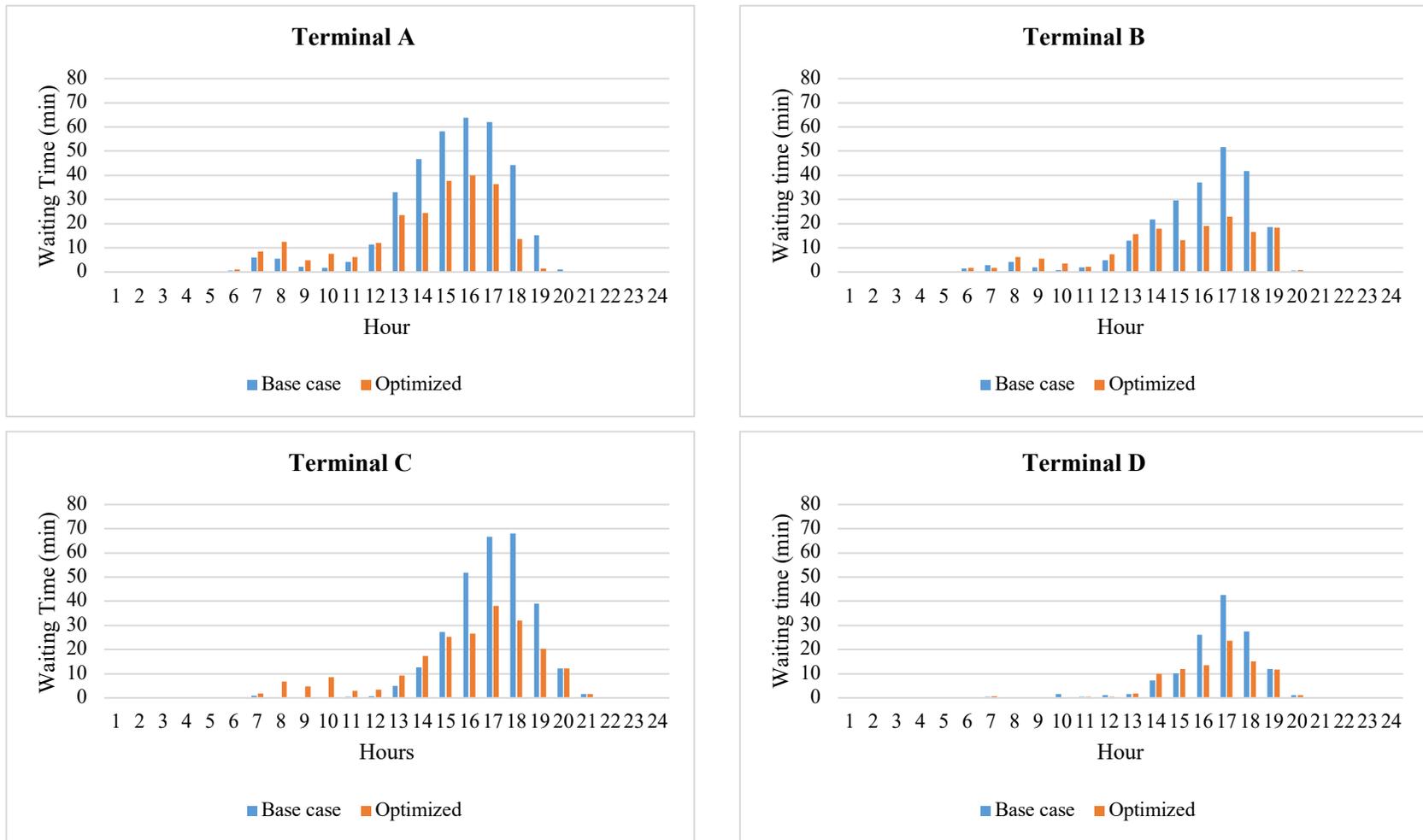

Figure 10: Waiting time comparison between the base case and optimized day

In Table 7, we show the cost and benefits from the perspective of all stakeholders in the system.

Table 7: Cost and benefit of stakeholders with TSMS (simulation of one working day)

| Stakeholder | Total gain [€] | Road Network [€] | | | | | Terminals [€] | | | |
|---|---|---|---|---|---|---|---|---|---|---|
| | | A15 | A4 | A16N | A16S | A29 | A | B | C | D |
| Terminal operator | 1025 | | | | | | 615 | 205 | 205 | 0 |
| Trucking companies | 15011 | | | | | | 4890 | 2905 | 4922 | 2294 |
| Trucking companies | -221.02* | | | | | | | | | |
| Traffic system | 1310 | 1904.3 | 250.16 | 170.27 | -1235.8 | 178.5 | | | | |
| # of trucks rescheduled | 231 (~10%) | | | | | | | | | |
| Computation time | 00:8:23 | | | | | | | | | |
| Waiting cost Optimality | 0.03 | | | | | | | | | |

*Disutility is unitless and should not be considered as monetary values

As we can see from Table 7, Trucking companies can gain 15011 euros on the day of the experiment experiencing less waiting time with only 10% of requests being shifted to off-peak period. Besides this gain, truckers can be more productive during the day if they don't have to wait at terminal gates. If we use 62 euros per hour as the cost of transporting a container (van der Meulen et al., 2020), dividing the total gain from less waiting time by 62 gives us the number of hours that truckers can be more productive along the day. In our case, the productivity gain of this solution is 241 hours.

The shifts in time slot requests, however, result in a total of 221.02 disutility for 231 trucks that have to reschedule their hinterland operations. Please note that this number is unitless and its magnitude cannot be compared with other costs in the system. This indicator shows how dissatisfied these carriers are from the perspective of their hinterland operation. Lower waiting time at gates resulted in a lower number of cranes being used by terminal operation in the hinterland side which leads to 1025 Euros per day (this gain will be higher for $\alpha > 1$). To investigate the relation between the gain of terminal operators and the disutility of carriers, we visualize the Pareto frontier of optimal solutions. Figure 11 shows that if we reduce the number of active cranes (reducing the cost of terminal operations) the TSMS system has to redistribute more requests to control truck arrival rates and consequently reduce waiting time at gates. This increases the disutility of carriers exponentially as they have to reschedule their hinterland operations. This graph can help decision-makers to tradeoff between the satisfaction of these two actors.

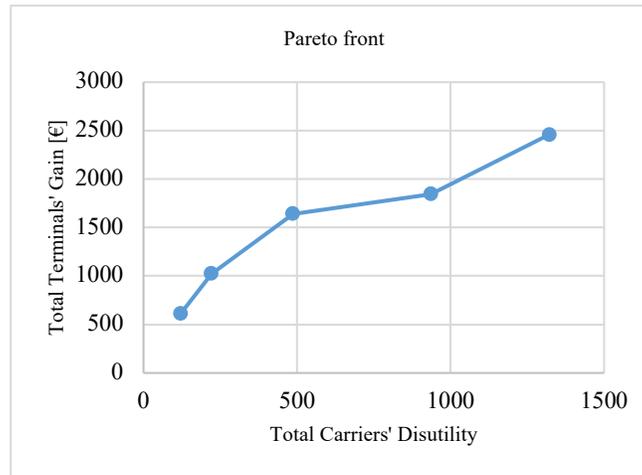

Figure 11: Terminal Gain Versus Carriers disutility in Pareto frontier of optimum solutions (only solutions with positive terminal gains are illustrated)

From the traffic perspective, we have small disturbances on the road networks between 15:00 and 18:00 on the day of the experiment and a large truck arrival rate for the same period. This is why TSMS has resulted in positive gains on 4 out of 5 motorways in the road networks by controlling truck arrival rates. The only motorway that has a negative gain (increased travel time along the day due to this operation) is A16-South which happened to have no congestion in the base case. The total gain of 1310 euros is relatively small and implies that TSMS not only prevents truck re-scheduling to deteriorate traffic conditions on road networks but also has a small impact on the improvement of traffic on the selected network. This impact however could be higher for larger networks with more severe congestion.

In Figure 12, we show the distribution of re-scheduled containers over commodity and container types. This shows which transport markets are more influenced by this system. As we can see that 34%, 27%, and 20% of all request that has been shifted belongs to Solid fuels, Chemical products, and Agricultural products respectively. General-purpose container, reefer, and chemical content containers have 74%, 15%, and 11% of the share of shifted requests. The reason is that these markets have lower disutility (see Table 5) for other times slots as compared to other markets which makes it possible for TSMS to shift their requests. These statistics only belong to the simulated day and more insight can be provided by TSMS if the system works over a longer period. Since this decision support system only rejects requests that have the minimum shift costs, such information or statistics can help decision-makers to identify the transport markets that have the potential to re-organize their hinterland activities and make a larger contribution to more efficient gate operation.

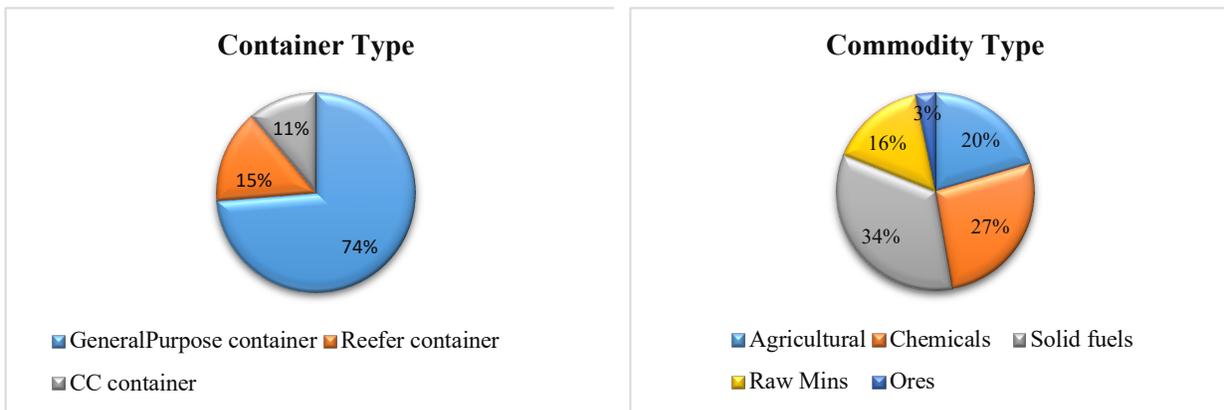

Figure 12: Share of shifted requests based on commodity and container types

## 6. Discussion

In this section, we summarize our main findings and discuss the potential of the proposed decision support system for time slot management in marine container terminals regarding its application for practice. We also discuss the limitations of the model and provide a set of recommendations for future research.

The results demonstrate the significance of integrative traffic and logistics modelling for use in terminal planning and decision support systems. This integration takes place considering the collective perspective of multiple actors which sheds light on the costs and benefits of the system. The data-driven truck scheduling model in the proposed TSMS provides insights into all these behaviours and preferences of truckers and hence can help decision-makers to interpret the result of this decision support system. For example, a decision-maker, by looking at the time slot preferences of all transport markets, can identify those in which their request is more frequently shifted to other time slots.

The costs and benefits of the stockholders involved in this system are not balanced. The gain for all stakeholders can be obtained at the cost of the disutility of a portion of carriers that are re-scheduled by the system. Our analysis shows that there is a benefit in the system for everyone but only truckers have to take action and bear the cost of re-organizing their schedules. From our simulation experiment based on the real case study, it is evident that the disutility of carriers increases if they have to re-schedule their tours to different pick-up times at terminal gates. Although taking into account the market preferences of carriers in TSMS minimizes the disutility of carriers, it still may put the success of the system at risk. Decision-makers may use the insight from TSMS to redistribute the collective benefit of the system to compensate the cost of carriers for equity. These incentives however require a gain-sharing mechanism to keep carriers motivated. The design of such a mechanism is beyond the scope of our study. However, we can picture it as tax reduction policies, giving priority to shifted trucks of specific markets that follow the TSMS recommendations, or defining incentive packages.

The first step towards designing such a system is to understand the monetary value of the disutilities. This requires a more in-depth analysis of hinterland pre-trip tour planning decisions like routing and scheduling of carriers. In this research, we only addressed hinterland scheduling decisions due to the available data. For calculating the monetary value of disutility of carries, however, a combined analysis of routing and scheduling decisions on a set of observed tours is required.

Finally, the impact of TSMS on road network traffic is highly dependent on the conditions of the road networks under study. The TSMS is not aimed at traffic management to help improve traffic conditions but can help to reduce the negative impact of the changes in port activities on traffic conditions. In addition, TSMS can provide predictions for traffic agencies. These can support the application of traffic management measures to reduce the possible negative impact of truck flows heading towards the hinterland or port. This highlights the importance of incorporating traffic modelling in the design of TSMS.

## 7. Conclusions

To deal with large waiting times, this paper presents a novel, centralized decision support system for time slot management at the marine container terminal. We explain the required steps toward the implementation of this system and evaluation of its performance under a simulation experiment, supported by real-world data. Our contribution is twofold.

Firstly, our method integrates logistics, freight and traffic modelling. It combines (1) advanced deep neural network structure for accurate truck demand predictions at terminal gates, (2) queuing modelling to simulate terminal gate operations, (3) utility maximization modelling to consider the disutility of truckers when assigning time slots, (4) a data-driven traffic model which can learn from day-to-day traffic to link logistics activities at the port to the dynamics of the traffic system and, finally (5) a nonlinear mixed-integer formulation to minimize multiple costs in the system. We conclude that the model can perform well in predicting truck demands ( up to 91% accuracy) at terminals gates, and predict traffic states on road networks with a 14% average error while linking these predictions to truck demands. The performance of the time slot management system indicates a significant reduction in waiting times at terminals' gates giving accurate suggestions to all the stakeholders. Secondly, the model provides new insights into the trade-off between supply (terminal operations) and demand (trucking operations) considering their impact on traffic at the port (terminal gates) and on-road networks in the hinterland. We infer the relation between the disutility of carriers and their tour planning decisions. By rescheduling a small portion of requests of trucks for time slots, the system obtains remarkable waiting time and productivity gains. We also find, however, that these gains come at the cost of carriers' processes. The system gives priority to trucks based on their preferences hence minimizing their costs by suggesting appropriate timeslot to drivers. We, therefore, conclude that the system coordinates well between multiple stakeholders in the port-hinterland ecosystem and provides an inclusive assessment of the costs\benefits of the system.

With respect to future research, we recommend including internal operations in the terminal model including the vessel discharge process. This could help the system to develop a broader view of the system's shadow costs. We also believe that insights into the monetary value of disutility of carriers, which can translate as truck scheduling cost, require more in-depth analysis of tours and trip chains in the hinterland. In addition to the stakeholders considered in this study, some of the shippers' and freight forwarders' decisions may have a direct or indirect impact on this system and therefore, requires further investigations. The application of such a decision support system can also be tested in digital twins of the port-hinterland ecosystem where policymakers can monitor day-to-day interaction between port and hinterland stakeholders.

## Acknowledgment
The funding body and partners of the research will be acknowledged following peer review.